\documentclass[letterpaper, 10 pt, conference]{IEEEtran}  

\IEEEoverridecommandlockouts                              

\title{\LARGE \bf
A Framework for Multisensory Foresight for Embodied Agents
}

\author{
\IEEEauthorblockN{Xiaohui Chen\textsuperscript{\textsection}, Ramtin Hosseini\textsuperscript{\textsection}, Karen Panetta, and Jivko Sinapov}
\IEEEauthorblockA{
\textit{Tufts University, MA, USA}\\
\{Xiaohui.Chen, Ramtin.Hosseini, Jivko.Sinapov\}@tufts.edu,  {Karen@ece.tufts.edu}} 
}

\usepackage{float}
\usepackage{amsmath,amssymb,amsfonts}
\usepackage{graphicx}
\usepackage{multirow}
\usepackage{caption}
\usepackage{subcaption}
\usepackage{hyperref}
\usepackage{makecell}
\usepackage{array}
\newcolumntype{P}[1]{>{\centering\arraybackslash}p{#1}}

\usepackage{color}

\begin{document}
\maketitle
\begingroup\renewcommand\thefootnote{\textsection}
\footnotetext{The first two authors contribute equally.}
\endgroup
\thispagestyle{empty}
\pagestyle{empty}


\begin{abstract}
Predicting future sensory states is crucial for learning agents such as robots, drones, and autonomous vehicles. In this paper, we couple multiple sensory modalities with exploratory actions and propose a predictive neural network architecture to address this problem. Most existing approaches rely on large, manually annotated datasets, or only use visual data as a single modality. In contrast, the unsupervised method presented here uses multi-modal perceptions for predicting future visual frames. As a result, the proposed model is more comprehensive and can better capture the spatio-temporal dynamics of the environment, leading to more accurate visual frame prediction. The other novelty of our framework is the use of sub-networks dedicated to anticipating future haptic, audio, and tactile signals. The framework was tested and validated with a dataset containing 4 sensory modalities (vision, haptic, audio, and tactile) on a humanoid robot performing 9 behaviors multiple times on a large set of objects. While the visual information is the dominant modality, utilizing the additional non-visual modalities improves the accuracy of predictions. 



\end{abstract}

\section{INTRODUCTION}

For humans and many animals, the ability to anticipate the future is a prerequisite for intelligent behavior. For robots, predicting the future values of sensors can assist object manipulation (\textit{e.g.} planning towards a desired sensory state), anomaly and failure detection (\textit{e.g.} by comparing predictions to observed values), and sensorimotor learning (\textit{e.g.} learning how sensors change as a result of the robot's actions). More generally, if a robot can predict the future values of sensors such as its cameras or haptic sensors, any perceptual routine that is used to process the robot's current sensory state would also be applicable for predicted sensory states. 

Early work in robotics focused on learning visual forward models that anticipate the future trajectories of objects manipulated by the robot as well as movements by the robot itself \cite{sigaud2007anticipatory}. More recently, methods have been developed to directly predict the future raw image frames that the robot would observe in its camera stream over the course of object manipulation \cite{finn2016unsupervised}. One limitation of existing methods is that they mostly deal solely in the visual domain. For many object manipulation tasks, however, other sensory modalities, such as haptic, audio and tactile, may be just as important. Non-visual sensory modalities can also help in situations where vision alone may be insufficient to resolve an ambiguity (\textit{e.g.} two objects may look identical but one may be much heavier than the other). Indeed research conducted in cognitive science \cite{wilcox2007multisensory, ernst2004merging} and robotics  \cite{bohg2017interactive,li2020review} has demonstrated the importance of using multiple (and often, non-visual) sensory modalities when learning about object properties and affordances.

Motivated by these findings, we present a deep learning methodology for {\it multisensory foresight} which uses feedback from multiple sensory modalities produced over the course of the robot's interaction with objects in its environment. We hypothesize that including more modalities can substantially improve prediction performance. To present and evaluate our proposed methodology, we used a publicly available dataset \cite{tatiya2019deep}, in which a robot performed 9 different types of exploratory behaviors (\textit{e.g.} \textit{push, press}, \textit{etc.}) on 100 objects multiple times. The dataset includes vision, haptic, audio, and vibrotactile sensory modalities. This paper introduces a modular deep neural network architecture that can take advantage of any modalities for performing the next-frame prediction task. Furthermore, we extend the model to predict the next frame for modalities other than vision, which leads to further improvements in the robot's prediction performance.

\begin{figure*}[t]
\begin{center}
\includegraphics[width=0.75\textwidth]{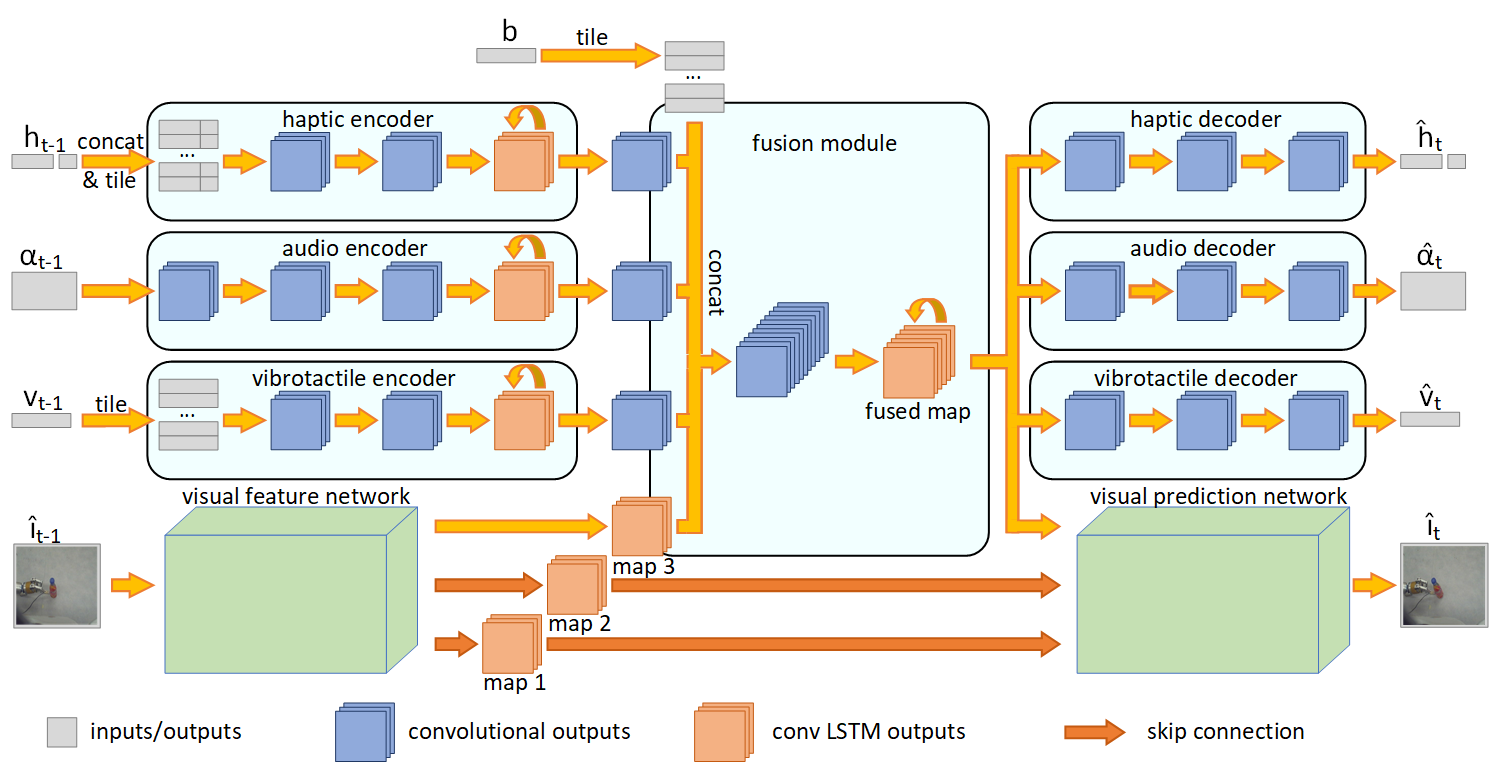}
\end{center}
\caption{The architecture of the proposed model, which consists of 4 feature encoders (left) and prediction heads (right) for 4 modalities, and 1 fusion module (middle) for merging representations of different modalities.}
\label{fig:arch}
\end{figure*}

\section{Related Work}
\textbf{Multi-modal perception.} A large volume of research has shown that perception can benefit by relating information
from multiple sources \cite{ngiam2011multimodal,7487176,li2020review,zhang2019leveraging,pastor2020bayesian}. To identify the semantics of objects (\textit{e.g.} empty, soft), visual information alone may not be adequate as the objects could be identical in the visual domain but different in other aspects (\textit{e.g.} material, internal state, compliance). To address this problem, several lines of research have focused on how robots can use non-visual sensory modalities of tasks that include grasping \cite{chitta2011tactile,zhang2019towards}, object recognition \cite{sinapov2011interactive,jin2019open,Gandhi-RSS-20}, object categorization \cite{tatiya2019deep,braud2020robot,tatiya2019sensorimotor} and language grounding \cite{chu2015robotic,amiri2018multi,benjamin2019improving,thomason2020jointly}. Inspired by these works, we propose an architecture that also uses multiple sensory modalities for the sensorimotor learning task of visual next-frame predication.

\textbf{Frame prediction.} This research aims to forecast future frames in video sequences. Early studies have focused on employing complex networks to directly generate pixel values (\textit{e.g.} \cite{yuen2010data}). However, these methods generally produce blurry predictions, as it is hard to model the distribution of image pixels, especially multiple steps into the future. Inspired by language modeling, Ranzato {\it et al.} \cite{ranzato2014video} applied a recurrent neural network to anticipate future frames. Srivastava {\it et al.} \cite{srivastava2015unsupervised} adapted LSTM model to capture pixel dynamics. Mathieu {\it et al.} \cite{mathieu2015deep} investigated different loss functions for sharper frame predictions. In another effort, Oh {\it et al.} \cite{oh2015action} proposed an action-conditional autoencoder network for Atari Games. Liang {\it et al.} \cite{liang2017dual} defined a dual motion Generative Adversarial Net (GAN). Recently a few approaches have solved the issue of blurriness of predictions multiple steps into the future \cite{wichers2018hierarchical,lee2018stochastic,babaeizadeh2017stochastic}. Despite the remarkable success, they have their own limitations. For example, \cite{wichers2018hierarchical} uses a hierarchical method which enables it to make sharper images for a longer period of time; however, it has the limitation that occasionally predictions disappear which constraints its applicability in safety critical settings. Two of the most successful models for frame prediction are PredNet \cite{lotter2016deep}, and the work introduced in \cite{finn2016unsupervised}. ConvLSTM units are essential building blocks of these two models. PredNet makes local predictions in each layer of the network and only passes deviations from the predictions to succeeding layers. The model presented in \cite{finn2016unsupervised} uses a pixel transformation function such as convolutional dynamic neural advection (CDNA) to predict motion distribution for the objects in videos. Despite the immense success, this model considers only one modality (vision) alongside state and action for forecasting future frames. In this paper, the proposed multi-modal network draws on the model architecture from \cite{finn2016unsupervised} for the vision prediction branch. By integrating several modalities to the network, the proposed model shows significant improvement in performance compared to the single-modality network.


\begin{figure*}[t]
     \centering
     \begin{subfigure}{0.495\textwidth}
         \centering
         \includegraphics[width=\textwidth]{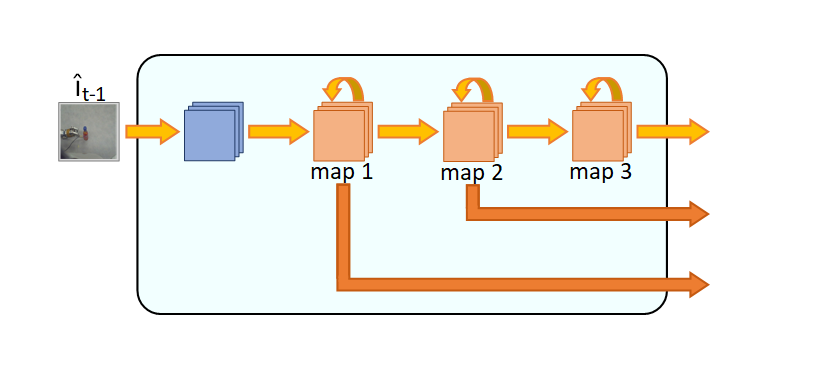}
         \caption{Visual Feature Network}
         \label{fig:visual feature network}
     \end{subfigure}
     \hfill
     \begin{subfigure}{0.495\textwidth}
         \centering
         \includegraphics[width=\textwidth]{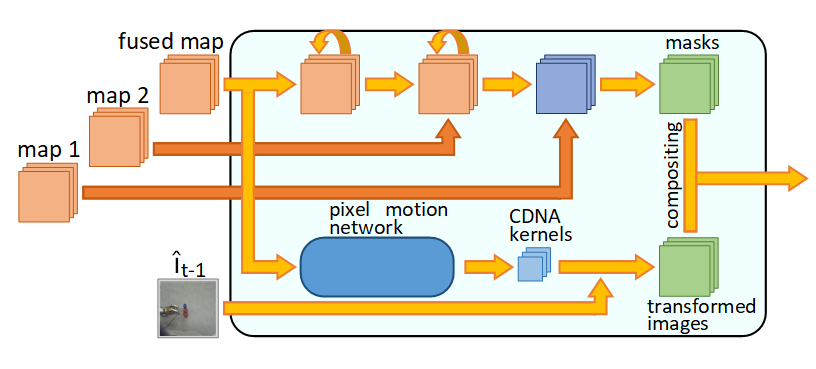}
         \caption{Visual Prediction Network}
         \label{fig:visual prediction network}
     \end{subfigure}
        \caption{Pipeline of The Visual Prediction Module, \ref{fig:visual feature network} shows the architecture of visual feature extractor, and \ref{fig:visual prediction network} shows the architecture of visual prediction network.}
        \label{fig:visual net}
\end{figure*}

\begin{figure*}[t]
     \centering
     \begin{subfigure}{0.32\textwidth}
         \centering
         \includegraphics[width=\textwidth]{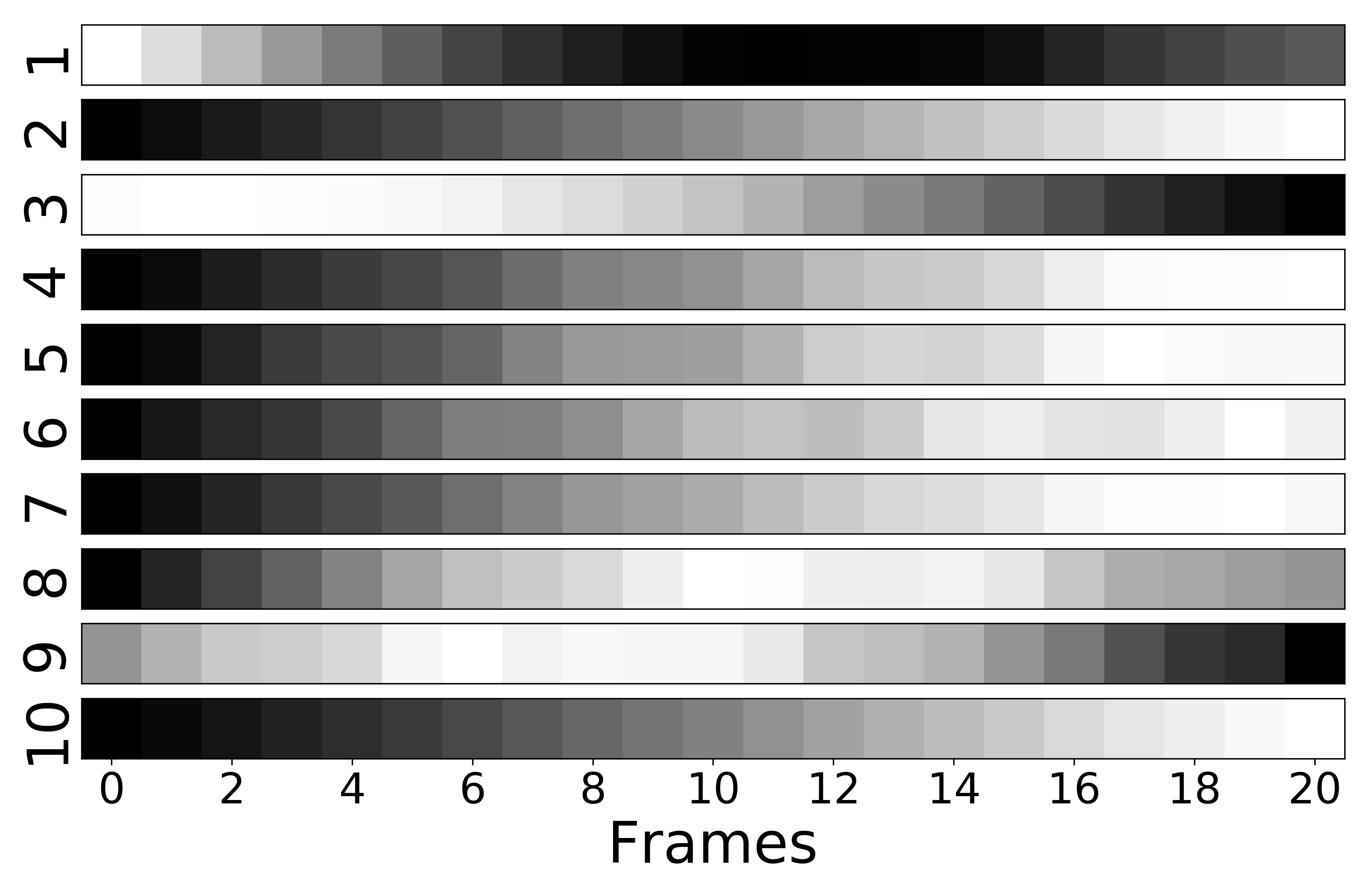}
         \caption{Haptic Features}
         \label{fig:hatpic_viz}
     \end{subfigure}
     \hfill
     \begin{subfigure}{0.32\textwidth}
         \centering
         \includegraphics[width=\textwidth]{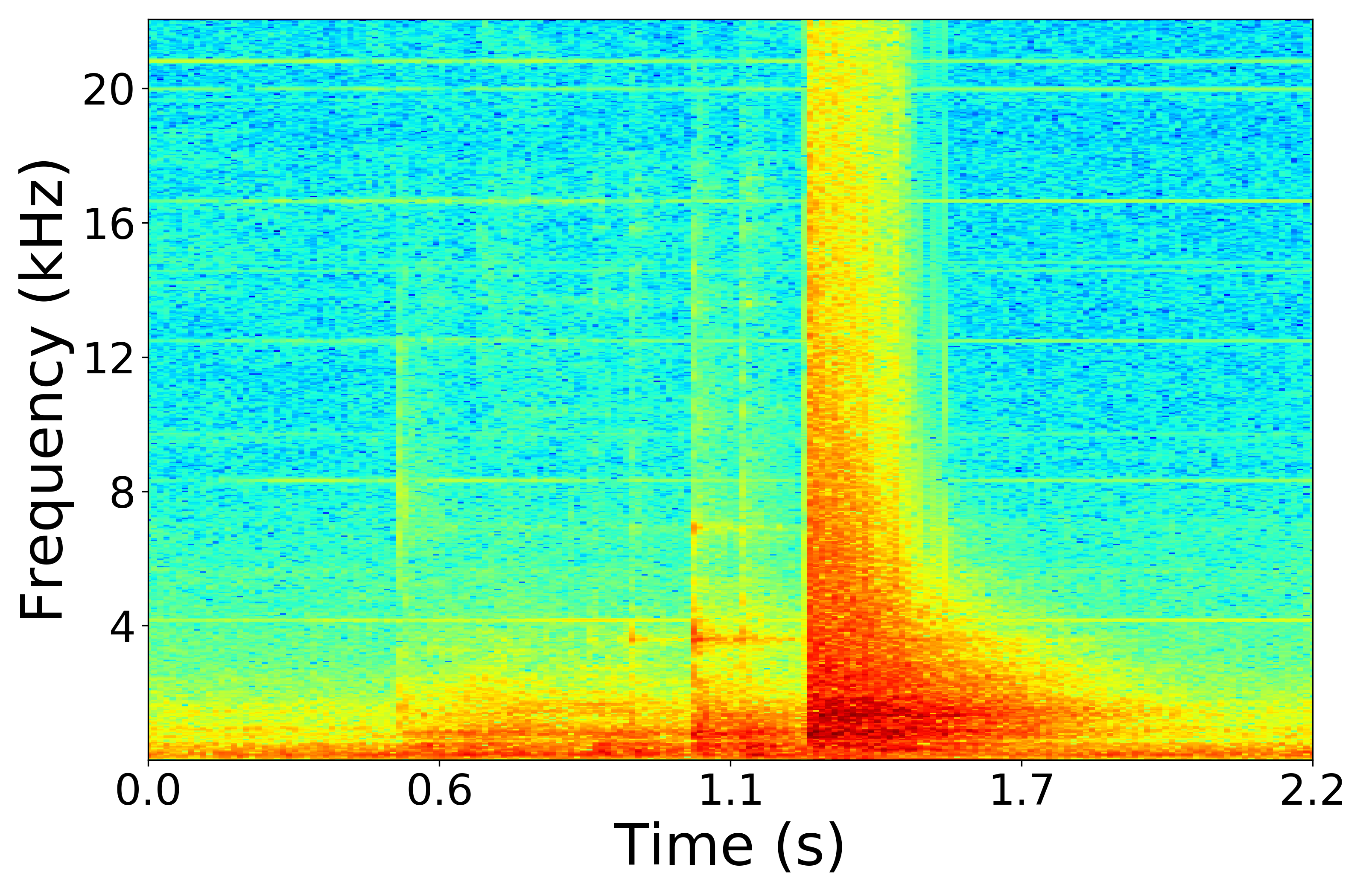}
         \caption{Audio Spectrogram}
         \label{fig:audio_viz}
     \end{subfigure}
     \hfill
     \begin{subfigure}{0.32\textwidth}
         \centering
         \includegraphics[width=\textwidth]{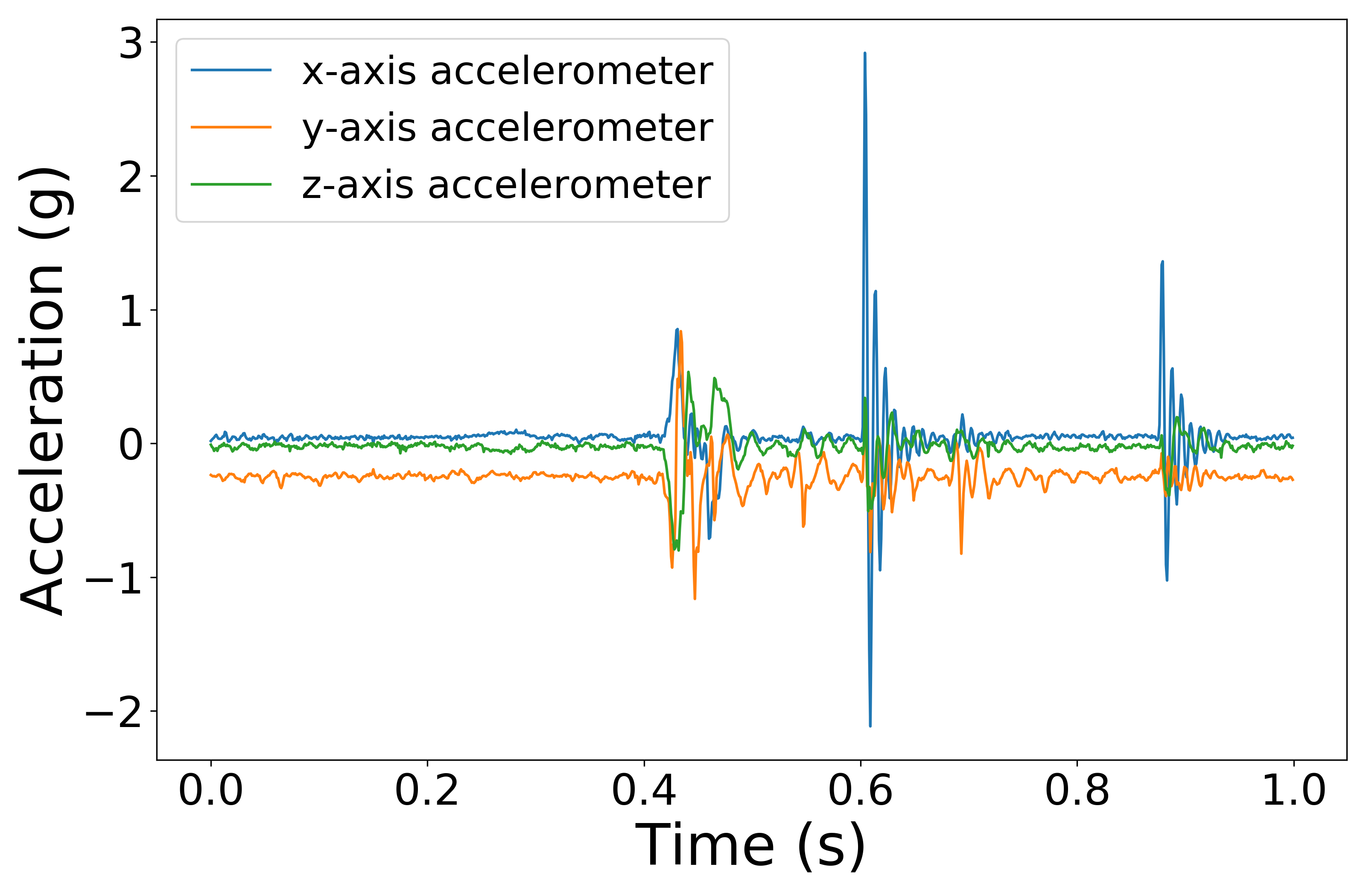}
         \caption{Vibro Accelerometer (3 Axes)}
         \label{fig:vibro_viz}
     \end{subfigure}
        \caption{Visualization of (a) haptic, (b) audio and (c) vibrotactile modalities when the robot \textit{drops} a bottle}
        \label{fig:Haptic&Audio&Vibro}
\end{figure*}
\begin{figure}[t]
 \centering
 \begin{subfigure}{0.49\textwidth}
     \centering
     \includegraphics[width=\textwidth]{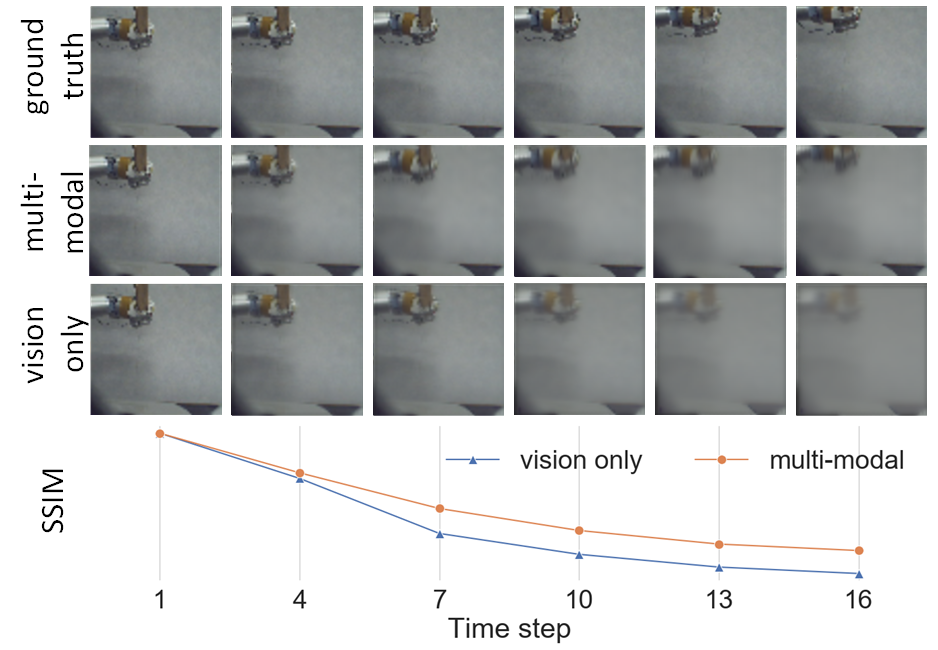}
     \caption{lift behavior}
     \label{fig:qualititive_lift}
 \end{subfigure}
 \hfill
 \begin{subfigure}{0.49\textwidth}
     \centering
     \includegraphics[width=\textwidth]{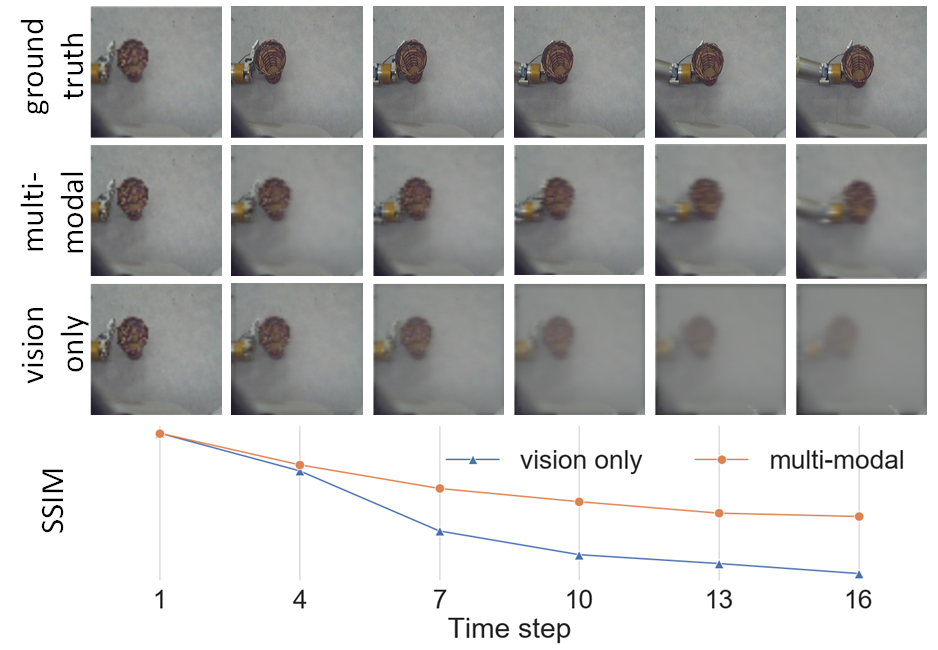}
     \caption{push behavior}
     \label{fig:qualititive_push}
 \end{subfigure}
\caption{Sharpness of predicted images, when the robot arm perform different behaviors (\ref{fig:qualititive_lift}: \textit{lift}, \ref{fig:qualititive_push}: \textit{push}).}
\label{fig:qualititative}
\end{figure} 

\section{Learning Methodology}
Next, we describe our framework for multisensory foresight, which uses multiple sensory modalities coupled with exploratory actions performed on objects by the robot. 


\subsection{Notation and problem formulation}
We used a dataset which contains $N$ samples $\{X^n\}$, where $n=1,2,\cdots, N$, and each sample $X^n$ is defined as a quadruple $X^n = \{\mathcal{I}^n, \mathcal{H}^n, \mathcal{A}^n,\mathcal{V}^n\}$. The quadruple is consisted of 4 kinds of sequential data collected by different sensors: 1) Visual data $\mathcal{I}^n = \{i_1^n, i_2^n, \cdots, i_T^n\}$; 2) Haptic data $\mathcal{H}^n = \{h_1^n, h_2^n, \cdots, h_T^n\}$; 3) Auditory data $\mathcal{A}^n = \{a_1^n, a_2^n, \cdots, a_T^n\}$; and 4) Vibrotactile data $\mathcal{V}^n = \{v_1^n, v_2^n, \cdots, v_T^n\}$. Different sensors of the robot execute at different frequency rate. As a result, with regard to our primary task which is to predict the following visual frames, all other modalities are processed to be synchronized to the visual data in terms of time step. To meet this end, for each time step, the modality data is defined as follows:

\begin{equation}
    \nonumber
    \begin{split}
    i_t^n\in\mathbb{R}^{M_w \times M_h \times M_c},
    h_t^n\in\mathbb{R}^{M_d\times M_d^\prime}\\
    a_t^n\in\mathbb{R}^{M_e\times M_e^\prime},
    v_t^n\in\mathbb{R}^{M_f\times M_f^\prime} \label{eq1}
    \end{split}
\end{equation}
where $M_w$, $M_h$ and $M_c$ are the width, height and the number of channels of each image respectively, $M_d$ is the number of robot joint-torque sensor readings, $M_e$ is the number of frequency bins in the audio spectrogram, and $M_f$ is the number of accelerometer readings. Moreover, $ M_d^\prime$, $ M_e^\prime$ and $ M_f^\prime$ are the number of in-frame time steps of haptic, auditory, and vibrotactile modalities respectively. 

The goal of the framework is to predict the future visual frames $\mathcal{I}_{>K}^n=\{i_{K+1}^n, \cdots, i_T^n\}$ given $K$ context frames $\mathcal{I}_{\leq K}^n=\{i_1^n, i_2^n, \cdots, i_{K}^n\}$ along with other modalities, where $K < T$. We also add a categorical feature $b^n\in\mathcal{B}$ indicating the type of behavior performed by the robot. While the main task is predicting subsequent frame images $\hat{i}_{t}^n$, where $t\in\{K+1,\cdots,T\}$, we introduce the concept of auxiliary tasks learning, which also predicts the next frames for haptic, audio and vibrotactile modalities which are denoted as $\hat{h}_{t}^n$, $\hat{a}_{t}^n$ and $\hat{v}_{t}^n$ respectively. Auxiliary tasks are expected to help find a stronger representation of how the modalities relate to one another through backpropagation, from which the main task might benefit. To this end, we define a highly abstracted autoregressive model $\mathcal{F}$:
\begin{equation}
     \hat{i}_{t}^n, \hat{h}_{t}^n, \hat{a}_{t}^n, \hat{v}_{t}^n = \mathcal{F}(\mathcal{I}_{\leq K}^n, \mathcal{H}_{<t}^n, \mathcal{A}_{<t}^n, \mathcal{V}_{<t}^n, b^n)
\end{equation}
where $\mathcal{H}_{<t}^n$, $\mathcal{A}_{<t}^n$, $\mathcal{V}_{<t}^n$ are the additional modality sequences prior to time step $t$. The model first learns how to extract high-level representations of each modality individually, then learns the interaction and combination of the 4 modality representations, and finally outputs the next frame prediction of each modality using the \textit{multi-heads} network. Next, we discuss the details about the model $\mathcal{F}$.

\subsection{Model architecture}
The proposed network architecture, shown in Figure \ref{fig:arch}, consists of 3 sub-modules: feature encoders, fusion module, and multi-modality prediction network.

\paragraph{Feature Encoders} Previous methods on next frame prediction relied mainly on the visual modality, while in our approach, inputs to the network are sequences of different modalities $\mathcal{I}$, $\mathcal{H}$, $\mathcal{A}$, $\mathcal{V}$. To efficiently integrate different modality features together, all modalities are mapped into $W \times H$ feature maps with different numbers of channels via their corresponding feature encoder. The feature encoder networks are composed of convolution, downsampling, and ConvLSTM modules with concatenation and tile operation.

For the visual modality, we employ stack ConvLSTMs (Figure \ref{fig:visual feature network}) to extract high-level vision features as well as spatio-temporal features. For the haptic modality, we spatially tile the concatenated joint signals and robot gripper pose across the feature map and feed it into the haptic-specific feature extraction network. For the audio and vibrotactile modalities, first we use Fast Fourier Transform (FFT) to compute a spectrogram, then employ convolutional layers and ConvLSTM layer to extract features.

\paragraph{Fusion Module} The fusion module contains one convolutional layer and one ConvLSTM layer with a concatenation operation, as illustrated in Figure \ref{fig:arch}. To further merge the modality features, it first integrates the lowest-dimensional activation maps given by each feature encoder into one latent feature map along the channel via concatenation operator, and feed it into the defined layers sequentially. The number of channels in the output feature map will be compressed into the same as of the visual input feature map, which in our work, channel size $64$, and $128$ are considered. The output feature map contains information extracted from all used modalities and will be further used to predict each modality in the next frame. Note that the number of chosen modalities can vary from 1 to 4, and the fusion module will automatically adapt the modality setting and output the integrated feature map with a fixed number of channels.

\paragraph{Multi-modal prediction head} The core of the model is learning the internal relation across different modalities, which consequently leads to increasing the performance of the main task (visual next-frame prediction). This is achieved by augmenting the auxiliary tasks. For each modality, there is a head that gets its input (fused feature map) from the fusion module, which integrates all the information and outputs the corresponding next frame modality.

For auxiliary task prediction heads, we directly reconstruct the next frame. Transposed convolutional layers are employed in each decoder, and the fused map is upsampled to be in the same dimension as the original input. For the visual prediction head (Figure \ref{fig:visual prediction network}), we use the idea of pixel transformation proposed in \cite{finn2016unsupervised, jaderberg2015spatial}, and perform two tasks instead of reconstructing the image directly. The first task is learning the pixel transformation parameters for each grouped object. The second task is performing an instance segmentation task that aims to group pixels by object. There are two branches for the visual prediction head. In the \textbf{object motion capture branch}, a motion prediction module called convolutional dynamic neural advection (CDNA) is used \cite{finn2016unsupervised}. The CDNA function computes new pixel values by applying multiple normalized convolution kernels to previous frames. CDNA is an object-centric motion prediction module, and as it is indicated in \cite{finn2016unsupervised}, the intuition behind it is that pixels form the same rigid entity move together. This module is expressed in the following equation:

\begin{equation}
\hat{J}_{t}(x,y)=\sum_{k\in(-k, k)}\sum_{l\in(-k, k)}\hat{m}(k,l)\hat{I}_{t-1}(x-k, y-l)
\end{equation}

\noindent where $k$ is the size of $\hat{m}$ convolution kernel, and ${\hat{J}}$ is a set of several transformations of the previous image. In the \textbf{instance segmentation branch}, skip connections are used to include the intermediate feature maps obtained from the visual encoder to the middle of the prediction head by directly concatenating them to restore the details learned in the low-level feature maps. This branch is responsible for applying masks to different objects. Finally, to obtain a single output image $\hat{I}_t$, the composition of predicted images should be modulated by a mask. 

\begin{equation}
\hat{I_t}=\sum_{c}\hat{J}_t^{(c)}\odot \Xi
\end{equation}

\noindent where $c$ represents the channel of the mask, and $\odot$ is the Hadamard product. The total loss function contains 4 components, each of which corresponds to the cost function for each modality. The cost function for each modality is weighted and is described below. $\mathcal{L}_{T}$ is the total loss:

\begin{equation}
     \mathcal{L}_{T} = \lambda_{i}\mathcal{L}_{i} + \lambda_{h}\mathcal{L}_{h}+ \lambda_{a}\mathcal{L}_{a}+ \lambda_{v}\mathcal{L}_{v}
\end{equation}
 
\noindent where in our work, the coefficient hyper-parameters are selected via grid search: $\lambda_{i}=1.0$, $\lambda_{h}=10^{-4}$, $\lambda_{a}=10^{-3}$ and $\lambda_{v}=10^{-4}$. We used mean square error (MSE) as the cost function for each modality.
 
\section{Experimental Results}
\label{sec:result}
We compare the proposed framework with the vision only model proposed in \cite{finn2016unsupervised} both quantitatively and qualitatively. To better investigate the robustness of the model, we provide two settings for experiments, which will be discussed in sections \ref{sec:exp1} and \ref{sec:exp2}. Furthermore, we discuss the effect of employing auxiliary training in section \ref{sec:exp3}.

\textbf{Implementation Details.} 
We make use of PyTorch \cite{paszke2017automatic} for GPU-based implementation\footnote{Code: \url{https://github.com/tufts-ai-robotics-group/mmvp/tree/main}}, set the number of context frames $K$ to 4, and evaluate the model performance for the following 16 predicted frames. For a few behaviors (\textit{grasp} and \textit{tap}), there are fewer frames in the dataset, only the following 6 frames are predicted. We employed ADAM optimizer \cite{kingma2014adam} with learning rate $lr=1e^{-3}$ to train the network for 30 epochs with batch size 32. For evaluation, we use Structural Similarity Index Measurement (SSIM) metrics to measure the visual prediction quality. Alternative metrics, such as Maximum Mean Discrepancy (MMD) \cite{gretton2012kernel} could be considered. We performed 5-fold cross-validation such that during each test, data from 80 objects was used for training and data from the remaining 20 objects was used for testing. 


\begin{figure}[t]
     \centering
     \begin{subfigure}{0.49\textwidth}
        \centering
        \includegraphics[width=\textwidth ]{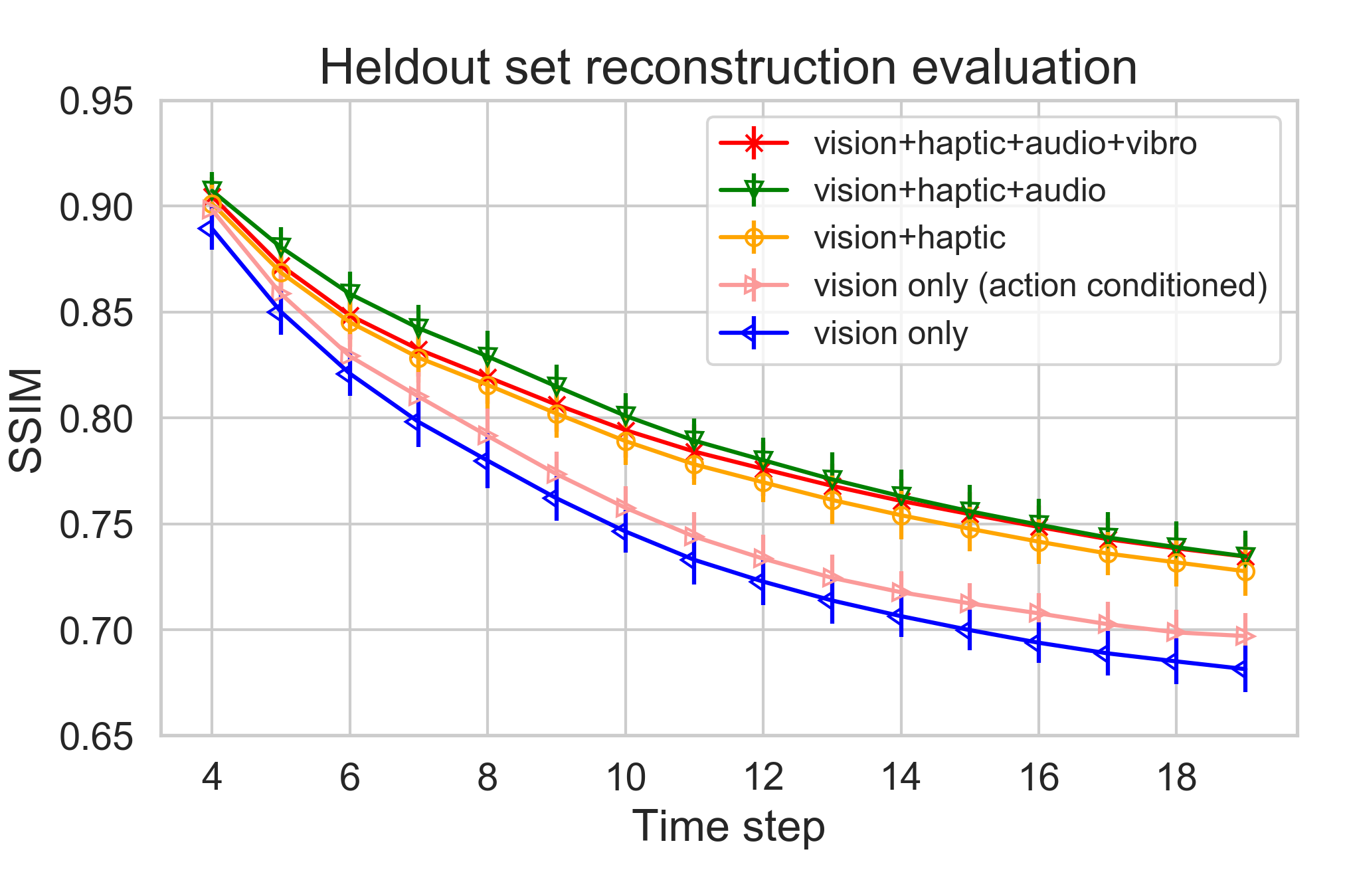}
        \caption{Ablation study on sensory inputs}
        \label{fig:all_behavior}
     \end{subfigure}
     \begin{subfigure}{0.49\textwidth}
        \centering
        \includegraphics[width=\textwidth ]{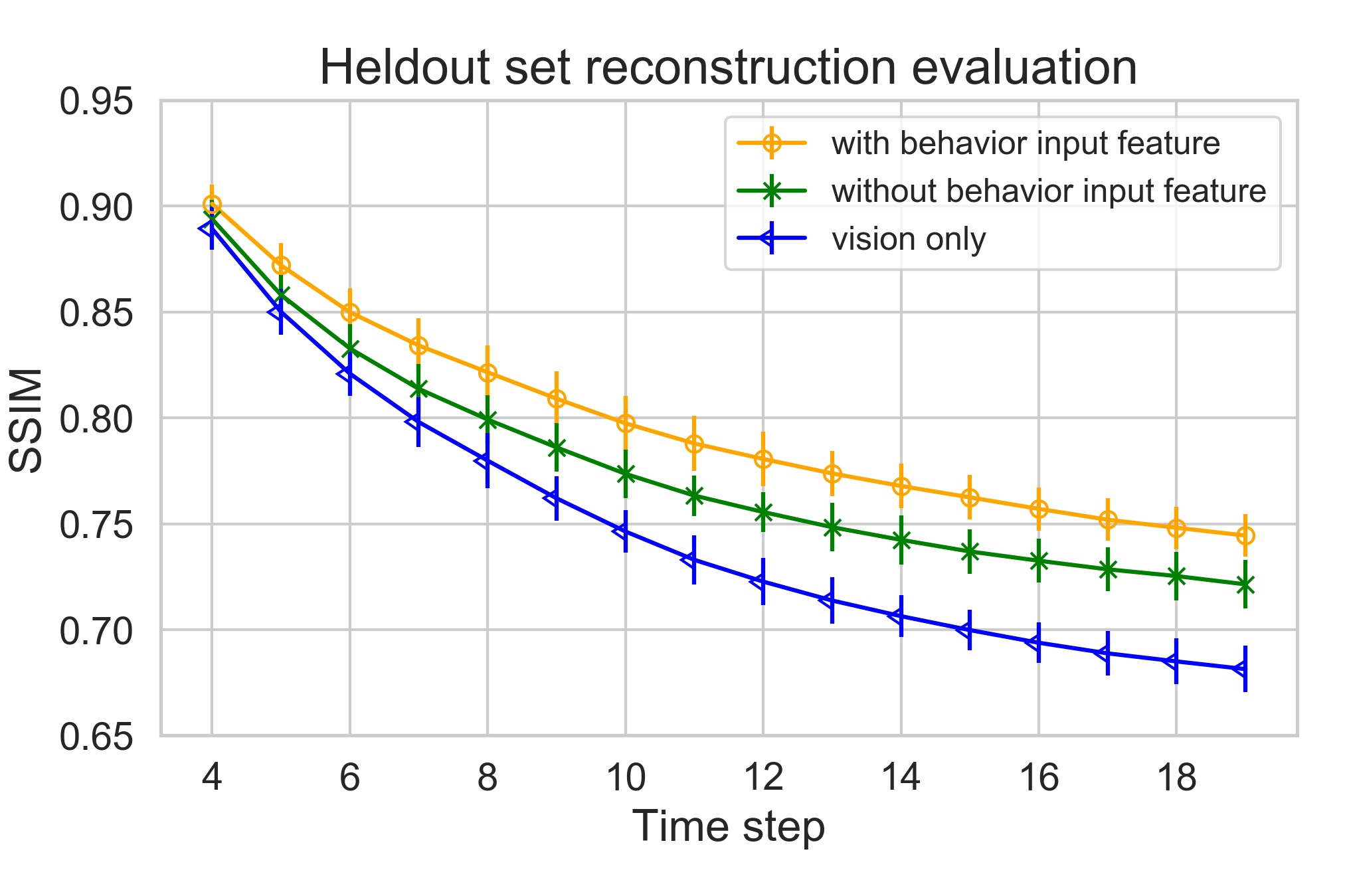}
        \caption{Ablation study on behavior input}
        \label{fig:ablation_on_behavior}
     \end{subfigure}
        \caption{Quantitative result evaluated with SSIM metric. Ablation studies on all behavior setting}
        \label{fig:all behavior}
\end{figure}


\begin{table}[t]
\begin{center}
\caption{Investigation of contribution of each modality to the improvement of model prediction}
\label{tb:detail ablation}
 \begin{tabular}{ c ||c |c| c|c} \hline\hline
avg. SSIM & haptic     & audio    & vibrotactile&   behavior   \\\hline
 0.771&\checkmark &          &                  &               \\ 
 0.773&\checkmark &\checkmark&                  &               \\ 
 0.767&\checkmark &          &\checkmark        &              \\ 
 0.769&           &\checkmark&                  &              \\ 
 0.756&           &          &\checkmark        &              \\ 
 0.770&           &\checkmark&\checkmark        &              \\ 
 0.776&\checkmark  &\checkmark&\checkmark        &          \\ 
 0.773&            &          &                  &     \checkmark         \\ 
\textbf{0.798}&\checkmark  &\checkmark&\checkmark        &\checkmark    \\ \hline\hline
\end{tabular}
\end{center}
\end{table}

\textbf{Dataset.} The dataset described in \cite{tatiya2020framework} is used to evaluate and compare the proposed network with the single-modal network. For collecting the dataset, an uppertorso humanoid robot with a 7-DOF arm manipulates 100 objects by executing 9 different exploratory behaviors (\textit{push, poke, press, shake, lift, drop, grasp, tap} and \textit{hold}) multiple times and records visual, haptic, auditory and vibrotactile sensory data. The visualization of different sensory modalities when the robot \textit{drops} a bottle is provided in Figure \ref{fig:Haptic&Audio&Vibro}. Figure \ref{fig:hatpic_viz} illustrates the torques of 7 joints of the robot and 3 end-effector positions over time. Figure \ref{fig:audio_viz} shows the spectrogram of the auditory data. We use the Fast Fourier Transform to convert the raw signal into a representation in the frequency domain. Figure \ref{fig:vibro_viz} shows the 3-axis accelerometer readings.


\begin{figure*}[t]
\begin{center}
\includegraphics[width=0.8\textwidth]{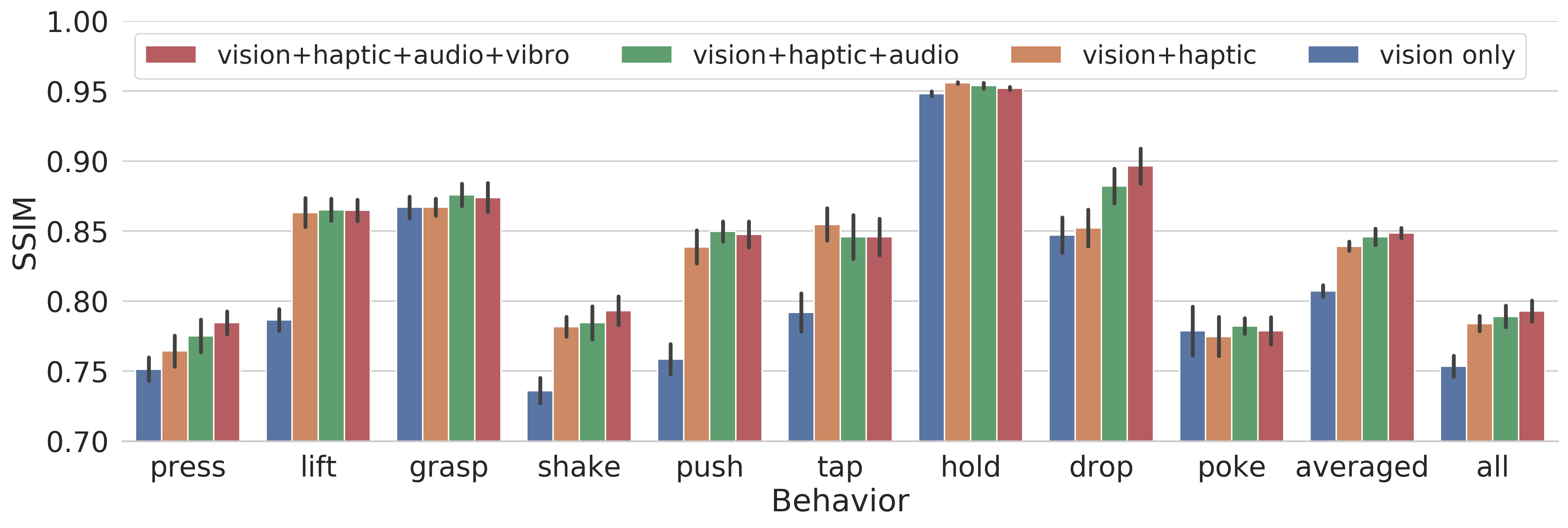}
\end{center}
\caption{Investigating the performance of different combinations of modalities per individual behavior}
\label{fig:sep_behave_group_bar}
\end{figure*}


\begin{table}[t]
\begin{center}
\caption{Modalities loss with and w/o auxiliary training, \textbf{aux} refers to auxiliary training and \textbf{no aux} refers to no auxiliary training.}
\begin{tabular}{p{2.9cm}||p{0.5cm}|p{0.76cm}||p{0.6cm}|p{0.6cm}|p{0.6cm}}
\hline\hline
\multirow{2}{*}{}         & \multicolumn{2}{P{1.5cm}||}{vision (SSIM)} & haptic (MSE) & audio (MSE) & vibro (MSE)\\ \cline{2-6} 
                          & aux             & no aux          & \multicolumn{3}{c}{aux}              \\ \hline
\centering{vision}        & \multicolumn{2}{P{1.5cm}||}{0.756}      &  \centering{-}          &  \centering{-}         &   \centering{-}    \tabularnewline \hline
\centering{vision+haptic}             & 0.785          & 0.764          & 0.282      &  \centering{-}         &    \centering{-}    \tabularnewline \hline
\centering{vision+haptic+audio}       & 0.796          & 0.791          & 0.246     & 0.042     &  \centering{-}    \tabularnewline \hline
\centering{vision+haptic+audio+vibro} & 0.798          & 0.795          & 0.244      & 0.041     & 0.739     \\ \hline\hline
\end{tabular}
\label{tb:other modalities}
\end{center}
\end{table}

\subsection{Training the Network with All Behaviors}
\label{sec:exp1}
The first experiment is to evaluate the framework in the all-behavior setting. Unlike the model in \cite{finn2016unsupervised}, which only uses one behavior (\textit{push}), in the presented work, we trained the model on data spanning all 9 exploratory behaviors and evaluated it on novel unseen objects that were not seen during training. In this setting, we first show an illustrative example which describes the qualitative results of using multi-modal perceptions and a vision-only model compared with ground truth. Then we quantitatively evaluate the model performance with regards to different numbers of used sensory modalities. Furthermore, we study the model's performance when the behavior type (\textit{e.g.} \textit{grasp} vs. \textit{push}) is added as a categorical feature to the network. Note that except when explicitly indicated, the behavior category feature is used as input for the experiments.

\textbf{Illustrative Example.}
Figure \ref{fig:qualititative} shows the qualitative reconstruction performance of the proposed method and vision-only model \cite{finn2016unsupervised} compared to ground-truth when the robot arm uses different behaviors (\textit{push, lift}) to interact with objects. We observe that predicted frames using multi-modal are much less blurry. Furthermore, this figure demonstrates that the proposed method better captures the motion and can localize the object appearance with more precision especially in multiple steps into the future (\textit{e.g.} see location of robot arm and the object for \textit{push} behavior, frame No. 16).

\textbf{Quantitative Reconstruction Performance.}  Figure \ref{fig:all_behavior} illustrates the performance of the network when integrated with different combinations of modalities compared to the vision-only method \cite{finn2016unsupervised}. The results show that utilizing the network with multi-modal perceptions substantially increases the performance of the predicted frames. Note the gap between vision only and any combination of multi-modal escalates for further future frames. Meanwhile, as expected, the quality of prediction in all models decreases over time as errors accumulate.
To avoid overfitting, we train the model with different channels in each layer and explore the effect of the model's size on the performance.
The baseline model explored in \cite{finn2016unsupervised} contains 12.5M parameters, based on which we extend other modality sub-networks and reached 13.6M parameters. The number of associated parameters for the additional modalities are much less for two reasons. First the dimensions of other modalities are smaller compared to the vision. Second a deeper network is used for the visual branch.
In another set of experiments, we investigate the effect of adding the behavior type as an input feature to the model. Figure \ref{fig:ablation_on_behavior} contrasts the model when it is trained with and without behavior. This figure shows the model performs better when the behavior is added as an input feature.

We demonstrate the contribution of each modality to the improvement of the model prediction via an ablation study. Table \ref{tb:detail ablation} shows the average SSIM over all time steps. The highest performance is obtained by integrating all modalities into the model. We also observe that in our dataset, haptic, audio, and behavior category share comparable contributions, while adding vibrotactile modality does not necessarily benefit the performance in this case, and sometimes it adds noise to the model which leads to performance degradation.

\subsection{Training Behavior-specific Models}
\label{sec:exp2}
We also investigate the performance of the model when trained and evaluated on an individual specific behavior. In this section, we ran the experiments with each behavior individually, yielding 9 models for each combination of modalities. We evaluate the performance of each model separately and also the averaged performance over all 9 behaviors. Furthermore, we compare the averaged performance to the model trained in section \ref{sec:exp1} under the same combination setting. Finally, we explore how each behavior model performs differently from the others and investigate how they benefit from the additional modalities. Figure \ref{fig:sep_behave_group_bar} shows the comparison between vision versus vision+haptic, vision+haptic+audio and vision+haptic+audio+vibrotactile for individual behaviors in terms of SSIM.

By comparing a different combination of modalities within each behavior, we observe that for 6 out of 9 behaviors, the model benefits from other modalities, especially, haptic. By contrasting the same modality setting across different behaviors, we notice that some behaviors (\textit{lift, grasp, hold} and \textit{drop}) pose an easier next-frame prediction challenge than others. We also observe that for tasks with discrete events (\textit{e.g. drop}), the audio and tactile modalities are very helpful for predicting future frames; however, for contact behaviors, the haptic modality is significantly more helpful than audio and tactile feedback. Furthermore, by integrating 9 separate models, we evaluate the averaged performance of the model (the 'averaged' column in figure \ref{fig:sep_behave_group_bar}). The averaged performance of the behavior-specific models is higher than that of the model trained simultaneously on all behaviors as described in Section \ref{sec:exp1}, shown in the rightmost column.

\subsection{Predicting Future Frames of Auxiliary Modalities}

\label{sec:exp3}
Another novelty of the proposed framework is predicting future frames of modalities other than vision. Predicting other modalities can sometimes be useful (\textit{e.g.} comparing the difference between predicted audio and the observed audio modality to identify abnormal events as they happen). In this subsection, we investigate the performance of these auxiliary tasks and whether learning them helps improve visual next-frame prediction. We evaluate vision modality prediction in two settings: \textbf{with auxiliary training} and \textbf{without auxiliary training} settings and assess the performance of the next-frame prediction model for the non-visual modalities under the \textbf{with auxiliary training} setting.

Table \ref{tb:other modalities} shows that auxiliary training of haptic modality enhances vision prediction while auxiliary training of audio and vibrotactile modalities does not necessarily contribute to improve visual next-frame prediction. Furthermore, this table shows that the audio modality contributes to the prediction of the haptic, while the vibrotactile modality seems to have little influence on predicting haptic and audio modalities.


\section{Conclusion and Future Work}
\label{sec:conclusion}
In this work, we developed a predictive framework incorporating multiple sensory modalities to help solve the next-frame prediction problem. Our experiments show that utilizing the network architecture with additional haptic, auditory, and tactile inputs achieves the best results compared to a state-of-the-art vision-only baseline. Furthermore, in this paper, we proposed the use of auxiliary tasks (predicting future haptic, audio, and vibrotactile signals) and showed that learning such tasks also improves visual next-frame prediction. 


One limitation of our framework is that it is trained on only one robot. Since different robots have different morphologies and different sensor suites, the learned knowledge cannot be directly used by another robot. An interesting avenue for future work is to extend transfer learning methodologies (e.g., \cite{tatiya2020haptic,tatiya2020framework}) as to enable a robot to bootstrap its sensorimotor learning process with knowledge learned by another robot. Another viable direction for future work is to integrate the multisensory next-frame prediction methodology described here with reinforcement learning methods for object manipulation tasks. 







\bibliographystyle{IEEEtran}
\bibliography{IEEEabrv,references}

\end{document}